\documentclass[conference]{IEEEtran}
\IEEEoverridecommandlockouts
\usepackage{cite}
\usepackage{amsmath,amssymb,amsfonts}
\usepackage{algorithmic}
\usepackage{graphicx}
\usepackage{textcomp}
\usepackage{xcolor}
\usepackage{array} 
\usepackage{hhline} 
\def\BibTeX{{\rm B\kern-.05em{\sc i\kern-.025em b}\kern-.08em
    T\kern-.1667em\lower.7ex\hbox{E}\kern-.125emX}}
\begin{document}

\title{A Comparative Study of Machine Learning Algorithms for Stock Price Prediction Using Insider Trading Data}

\author{\IEEEauthorblockN{Amitabh Chakravorty, Nelly Elsayed}
\IEEEauthorblockA{\textit{School of Information Technology} \\
\textit{University of Cincinnati}\\
Cincinnati, OH, USA \\
chakraa4@mail.uc.edu, elsayeny@ucmail.uc.edu}
}

\maketitle

\begin{abstract}
The research paper empirically investigates several machine learning algorithms to forecast stock prices depending on insider trading information. Insider trading offers special insights into market sentiment, pointing to upcoming changes in stock prices. This study examines the effectiveness of algorithms like decision trees, random forests, support vector machines (SVM) with different kernels, and K-Means Clustering using a dataset of Tesla stock transactions. Examining past data from April 2020 to March 2023, this study focuses on how well these algorithms identify trends and forecast stock price fluctuations. The paper uses Recursive Feature Elimination (RFE) and feature importance analysis to optimize the feature set and, hence, increase prediction accuracy. While it requires substantially greater processing time than other models, SVM with the Radial Basis Function (RBF) kernel displays the best accuracy. This paper highlights the trade-offs between accuracy and efficiency in machine learning models and proposes the possibility of pooling multiple data sources to raise prediction performance. The results of this paper aim to help financial analysts and investors in choosing strong algorithms to optimize investment strategies.
\end{abstract}


\section{Introduction}
For a long time, financial analysts and investors have been interested in the subject of insider trading~\cite{b3}. Insider Trading refers to buying or selling securities of a publicly traded company by employees who possess confidential information about that company that has not been made public~\cite{b8,b10}. Insider trading is regulated by laws, and when insiders deal in equities in accordance with those regulations, their acts are regarded as legal~\cite{b8}. Data on insider trading is a major tool for predicting stock prices as it reflects the actions and expectations of those with the most knowledge of the financial condition and future changes of a company~\cite{b18}. 

Exploring trends in insider trading data might help determine whether insiders have a favorable or unfavorable outlook on the future of the company~\cite{b7}. This type of data offers a distinctive perspective on the mood within the company, which might suggest forthcoming stock price changes before they show in the market~\cite{b7}. For instance, if a significant number of executives are purchasing their company’s stocks, it can be a sign that they are optimistic about the future performance of the business, which could raise stock prices. On the other hand, if insiders are selling their shares, it can be a sign that they are less optimistic about the future prospects of the business, which could result in a decline in the stock prices of that company. Insider trading may also take place before significant market-moving events like mergers, acquisitions, financial disclosures or leadership changes~\cite{b19}. In light of this, analyzing insider trading data can deliver investors and financial analysts valuable details to help them decide whether to purchase or sell certain stocks.

By using data-driven algorithms to automate prediction, Machine Learning (ML) can transform the process of predicting stock prices~\cite{b5,b4,b3}. Machine Learning algorithms are AI programs that change and evolve in response to the data they process to deliver preset results~\cite{b15}. In essence, they are mathematical constructs that have the ability to learn from the statistics they are fed, which is frequently referred to as training data~\cite{b15}. The ability of these algorithms to learn from historical data and identify patterns that are invisible to the human eyes is their prime feature. By analyzing the data of insider trading activities, these algorithms can identify key factors that impact stock prices and use that detail to make accurate predictions.

It is vital to note that these algorithms are not a silver bullet for predicting stock prices. Many factors can influence stock prices, including geopolitical events, economic indicators, and changes in consumer behaviors, which may not be captured in the stock trading datasets used to train the algorithms. Additionally, the accuracy of machine learning predictions depends heavily on the quality of the data used to train the algorithms~\cite{b15}. If the data is incomplete or inaccurate, the machine learning algorithms might yield inaccurate predictions.

This research paper aims to investigate, assess and analyze the performance of popular machine learning algorithms in predicting stock prices from insider trading data. In addition, this paper aims to identify the areas where further improvements are required to boost the accuracy of predictions. The findings of this research can help investors and financial analysts get an idea of the performance levels of popular machine learning algorithms for stock market predictions and, hence, help them select and use the optimal algorithm to decide on better investments.

\section{Methodology}

The empirical study outlined in this paper involved acquiring a suitable dataset of insider. Then the data processing and a feature selection was conducted to identify the most relevant features in the dataset for predicting stock prices. This involved using a statistical technique called feature importance analysis. The machine learning algorithms were trained using the dataset and run on the test dataset to predict stock prices by using the selected features of the insider trading dataset. 
This empirical study employed the Decision Trees, Random Forests, and Support Vector Machines (SVM) as three different supervised machine learning methods to predict stock prices from insider trading datasets. SVM algorithms use a group of mathematical functions which are known as kernels~\cite{b11}. The kernels of SVM used in this research were linear, polynomial, and radial basis functions (RBF). These algorithms were selected due to their capability to handle high dimensional data, nonlinear correlations between characteristics and deliver findings that are easy to understand. K-Means Clustering, an unsupervised machine learning tool~\cite{b13}, was also utilized to predict the prices of stock by finding patterns and abnormalities in the data. 

The Decision Tree method is used to estimate a target function that has discrete values and represents the learned function as a decision tree structure~\cite{b1}. The decision tree classifies the instances as they are processed by arranging them according to the feature values from root to leaf nodes~\cite{b1}.

The Random Forest is an ensemble learning technique employed in both classification and regression tasks. The method uses a bagging approach to create a collection of decision trees using a randomly selected subset of the available data~\cite{b2}. The output from each decision tree is then merged to arrive at the final classification or regression decision~\cite{b2}.
The Support Vector Machine (SVM) is a supervised learning algorithm capable of performing both classification and regression tasks. When each data point is represented as a point with n coordinates equal to the number of features in the dataset, this method computes margins in a high-dimensional space~\cite{b2}. Finding a hyperplane that maximally divides the training data into discrete classes is the objective of this algorithm. SVMs accomplish this by locating the hyperplane that is the furthest away from the nearest points of data in both classes. In other words, the hyperplane is chosen to maximize the distance between the closest data points of each class, providing an optimal separation boundary~\cite{b2}.

The K-means is a frequently used unsupervised learning algorithm for cluster analysis. Its main goal is to divide a set of n observations into a set of k clusters, with each observation being assigned to the cluster whose mean is closest to it and acting as a prototype for the cluster~\cite{b2}. The mean of the observations in that particular cluster serves as the cluster's center~\cite{b2}.

\begin{figure*}[htbp]
	\centering
	\includegraphics[width=11cm, height= 7cm]{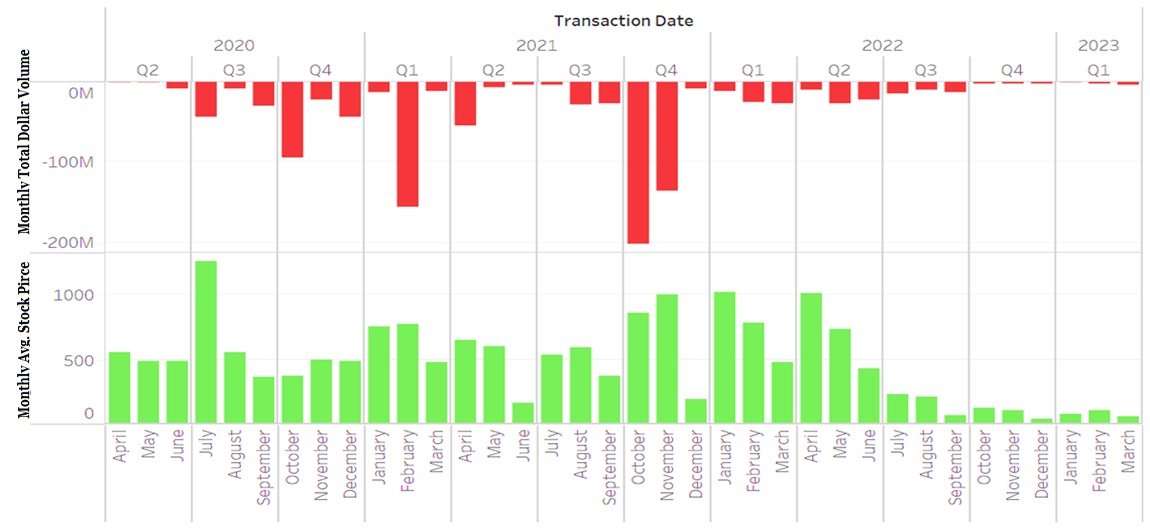}
	\caption{Bar graphs of Tesla’s monthly average stock prices and monthly total dollar volume traded.}
	\label{fig1}
\end{figure*}

\section{EXPERIMENTAL SETUP}
The dataset consists of historical trading data of Tesla from April 2020 to March 2023. The insider trading data of different popular public companies, viz., Goldman Sachs~\cite{b12}, Facebook~\cite{b12}, Google~\cite{b12}, Tesla~\cite{b12}, Twitter~\cite{b12}, and Gamestop~\cite{b12} were carefully examined. After observing the insider trading data of these popular companies, it was found that Tesla had the highest number of insider transactions from April 2020 to March 2023, and hence, Tesla's stock was chosen to be analyzed to predict its prices. The dataset includes the variables, viz., Change, filingDate, Name, Shares, symbol, transactionID, transactionCode, transactionDate, and Price on the trading activities of Tesla. The Change variable refers to the change in the insider's portfolio after a certain transaction. The dataset consists of 1997 data samples. The dataset was preprocessed to remove any missing or inconsistent data, normalize the features, and transform the data into a suitable format for analysis. The irrelevant rows of data and irrelevant columns, viz., transactionID, symbol, and transaction code, were removed from the dataset, which was obtained after running the Python script written to scrape the insider trading transactions data on the web.  

The trading data of Tesla was mined from a financial website called Finnhub.io. Finnhub.io offers its users access to different forms of stock trading data~\cite{b12}. The insider transactions API (Application Programming Interface) request of Finnhub.io was used in this research. A single request allows mining the trading data for a time period of 12 consecutive months, and so three requests were written to mine the trading data of the previous 36 consecutive months in the Python script for data mining. The string data was converted into a data frame, which is a dataset organized into a two-dimensional table of rows and columns, to perform the data manipulation and analysis better. From the variables in the mined dataset, two more attributes were derived in this research, which would work as vital indicators of stock prices. The derived variables are Dollar Volume and Type.

The relevance of the derived variables in terms of predicting stock prices is the dollar volume and the type. The Dollar Volume variable is the total dollar amount of a transaction conducted by a certain executive at a certain time. Increases in a transaction's dollar value that are out of the ordinary may be a sign of insider trading activity. This is due to the possibility that insiders who have access to private information may exploit it to their advantage by purchasing or disposing of the relevant security, which might result in a sharp rise or fall in the Dollar Volume variable. The Type variable demonstrates if the transaction was a buy, sale, or gift. It is crucial to know the type of transaction to analyze the insider trading data. All the gift transactions present in the dataset indicate employee benefits given to the insiders of the company. If the transaction involved a gift, insider information was not involved. The only transactions that can be examined to see if they were insider trading are buys and sales. The dataset was cleaned by dropping irrelevant data rows of gift transactions to improve the experiment procedure since only buys and sales influence the stock prices.

The dataset was divided into training and testing sets, with 70 percent of the data used for training and the remaining 30 percent used for testing. The machine learning algorithms were first trained using the training set. After training, the test sets were fed to the algorithms to get the prediction results. The effectiveness of the various machine learning algorithms was examined using the evaluation criteria of accuracy~\cite{b14} and the time required to deliver results in order to identify the best-performing algorithm. 

In order to find the most important traits of insider trading data to predict stock prices, a feature importance analysis was also carried out prior to training the machine learning algorithms. In order to choose the most useful features, the Recursive Feature Elimination (RFE) method was applied to the research dataset.

The RFE method works by recursively removing features from the dataset and fitting the model on the remaining features until the desired number of features is reached~\cite{b16}. Each of the four algorithms used in this experiment was first trained on the entire dataset, and then the feature importance of each feature was computed. The least important variable or feature was removed, and the algorithms were trained again on the remaining variables. This process was repeated until the desired number of features was reached and until the performances of the algorithms did not improve significantly. The desired number of variables for this research was four.

\begin{table}[htbp]
	\centering
	\caption{Accuracies and consumed times of the ML algorithms sorted in descending order.}
	\label{tab:ml_comparison}
	\begin{tabular}{|l|c|c|}
		\hline
		\textbf{ML Algorithm} & \textbf{Prediction Accuracy} & \textbf{Time} \\ \hline
		SVM RBF & 88\% & 28 minutes \\ \hline
		Random Forest & 83\% & 18 minutes \\ \hline
		SVM Polynomial & 81\% & 12 minutes \\ \hline
		SVM Linear & 77\% & 8 minutes \\ \hline
		K-Means Clustering & 73\% & 7 minutes \\ \hline
		Decision Tree & 68\% & 1 minute \\ \hline
	\end{tabular}
\end{table}

\begin{figure*}[htbp]
	\centering
	\includegraphics[width=11cm, height= 6 cm]{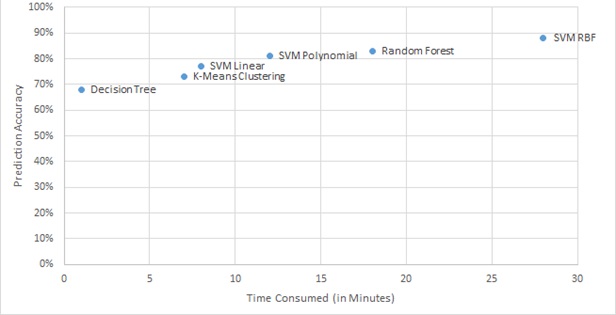}
	\caption{Scatterplot of accuracies versus consumed times of the ML algorithms.}
	\label{fig2}
\end{figure*}

\section{Results}
After carrying out the recursive feature elimination method, it was determined that the most pertinent variables of the dataset to determine the stock prices were Shares, Transaction Date, Dollar Volume, and Type. Among these variables, the Dollar Volume variable had the highest relevance.

Figure~\ref{fig1} shows the high relevance of the Dollar Volume variable to predict stock prices, bar graphs of the average price of the Tesla stocks, and the total dollar volume traded by the executives of Tesla for each month from April 2020 to March 2023.

According to Table~\ref{tab:ml_comparison} and Figure~\ref{fig2} given below, the highest accuracy rate of predicting stock prices using a machine learning algorithm was obtained at 88 percent, and the lowest was obtained at 68 percent. This experimental research states that Support Vector Machines (SVMs) with the Radial Basis Function (RBF) kernel and Random Forest are the most effective models for predicting stock prices using insider trading data. Nonetheless, it took the longest time for these models to predict the results. On the other hand, the fastest model to predict the stock prices was the Decision Tree, but it had the lowest accuracy rate. According to the results obtained from this research, it can be comprehended that the longer the machine learning algorithm takes to predict the results, the higher the accuracy rate of that algorithm. 

The computational analyses for this research were conducted on a Lenovo ThinkPad E14 Gen 2, equipped with an AMD Ryzen 5 4500U processor, 16 GB of RAM, and a 512 GB SSD, running Windows 10 Professional, 64-bit. The primary programming environment was Python 3.8.5, enhanced by several machine learning libraries. The library scikit-learn was predominantly employed for the machine learning tasks. Data manipulation and numerical operations were managed using Pandas and NumPy libraries.
The accuracies of the results predicted and the time taken to deliver the results by the algorithms are given in Table~\ref{tab:ml_comparison}.

Support Vector Machine algorithm with the RBF kernel took the longest time to deliver the predictions compared to the other machine learning algorithms due to its computational complexity. The Radial Basis Function kernel is a non-linear kernel function that allows SVM to model complex decision boundaries~\cite{b6}. This kernel function has a large number of hyperparameters, like the regularization parameter and the gamma parameter~\cite{b6}. These hyperparameters can significantly impact the model's accuracy rate and computational complexity. In addition, SVM with RBF kernel requires solving a quadratic optimization problem for each training example, which can be computationally heavy for large datasets. Therefore, SVM with RBF kernel can take more time than the simpler machine learning algorithms like decision trees, SVM with Linear kernel, and K-Means Clustering. Nonetheless, this increased computational complexity was worth it since SVM with RBF kernel delivered the best accuracy rates in the prediction of the stock prices among all the algorithms assessed in this research.

\section{Conclusion}
Insider trading details illustrate significant indications about market sentiment and assist investors in making wise financial decisions. In this experimental research, the use of machine learning algorithms to predict stock prices using insider trading data was explored.

The results of this research indicate that machine learning algorithms can be an effective tool in predicting stock prices. The four popular algorithms that were evaluated in this research are decision trees, random forests, Support Vector Machines, and K-means clustering. The evaluation results specified that SVM with Radial Basis Function kernel outperformed the other models in terms of accuracy. SVM is an effective method for managing massive datasets and has been extensively utilized in many sectors of finance. The results imply that SVM can be a useful tool for financial professionals and investors to forecast stock prices.

Nonetheless, a sizable amount of data is needed to apply machine learning algorithms in stock price prediction. It may not be possible to forecast stock prices accurately using just insider trading data. Additional data sources, like news stories, financial reports, and social media, can also give insightful details about the state of the stock market and help improve the accuracy of stock price predictions. In order to make better investment judgments, financial analysts and investors must consider utilizing multiple sources of data.

It is essential to accept the fact that machine learning algorithms cannot be used as the sole means of making investment decisions. The stock market is a dynamic and complicated system which is impacted by a variety of social, political, and economic factors. No algorithm can guarantee a precise prediction of the ways these factors would affect stock prices. When making investment decisions, machine learning algorithms can be employed as a tool rather than the exclusive method to predict stock prices.



\bibliographystyle{IEEEtran}
\bibliography{references}

\begin{thebibliography}{10}
\providecommand{\url}[1]{#1}
\csname url@samestyle\endcsname
\providecommand{\newblock}{\relax}
\providecommand{\bibinfo}[2]{#2}
\providecommand{\BIBentrySTDinterwordspacing}{\spaceskip=0pt\relax}
\providecommand{\BIBentryALTinterwordstretchfactor}{4}
\providecommand{\BIBentryALTinterwordspacing}{\spaceskip=\fontdimen2\font plus
\BIBentryALTinterwordstretchfactor\fontdimen3\font minus
  \fontdimen4\font\relax}
\providecommand{\BIBforeignlanguage}[2]{{%
\expandafter\ifx\csname l@#1\endcsname\relax
\typeout{** WARNING: IEEEtran.bst: No hyphenation pattern has been}%
\typeout{** loaded for the language `#1'. Using the pattern for}%
\typeout{** the default language instead.}%
\else
\language=\csname l@#1\endcsname
\fi
#2}}
\providecommand{\BIBdecl}{\relax}
\BIBdecl

\bibitem{b3}
M.~M. Kumbure, C.~Lohrmann, P.~Luukka, and J.~Porras, ``Machine learning
  techniques and data for stock market forecasting: A literature review,''
  \emph{Expert Systems with Applications}, vol. 197, p. 116659, Jul. 2022.

\bibitem{b8}
U.~Bhattacharya and H.~Daouk, ``The world price of insider trading,''
  \emph{SSRN Electronic Journal}, vol.~57, no.~1, pp. 75--108, Dec. 2000.

\bibitem{b10}
C.~T. Lundblad, Z.~Yang, and Q.~Zhang, ``Detecting insider trading in the era
  of big data and machine learning,'' \emph{SSRN Electronic Journal}, Sep.
  2022.

\bibitem{b7}
P.~Mazzarisi, A.~Ravagnani, P.~Deriu, F.~Lillo, F.~Medda, and A.~Russo, ``A
  machine learning approach to support decision in insider trading detection,''
  \emph{SSRN Electronic Journal}, Dec. 2022.

\bibitem{b5}
S.~Tiwari, A.~Bharadwaj, and S.~Gupta, ``Stock price prediction using data
  analytics,'' in \emph{2017 International Conference on Advances in Computing,
  Communication and Control (ICAC3)}, Mumbai, India, 2017, pp. 1--5.

\bibitem{b4}
K.~V. Kumar and R.~Anitha, ``A detailed survey to forecast the stock prices by
  applying machine learning predictive models and artificial intelligence
  techniques,'' in \emph{2022 International Conference on Computing,
  Communication, Security and Intelligent Systems (IC3SIS)}, Kochi, India,
  2022, pp. 1--6.

\bibitem{b15}
\BIBentryALTinterwordspacing
Arm, ``What are machine learning algorithms?'' Arm Glossary, 2023, accessed:
  17-Mar-2023. [Online]. Available:
  \url{https://www.arm.com/glossary/machine-learning-algorithms}
\BIBentrySTDinterwordspacing

\bibitem{b11}
\BIBentryALTinterwordspacing
S.~Awasthi, ``Seven most popular svm kernels,'' Dataaspirant, 2023, accessed:
  14-Mar-2023. [Online]. Available: \url{https://dataaspirant.com/svm-kernels/}
\BIBentrySTDinterwordspacing

\bibitem{b13}
\BIBentryALTinterwordspacing
Javatpoint, ``K-means clustering algorithm,'' 2023, accessed: 11-Apr-2023.
  [Online]. Available:
  \url{https://www.javatpoint.com/k-means-clustering-algorithm-in-machine-learning}
\BIBentrySTDinterwordspacing

\bibitem{b1}
R.~Shokri, M.~Stronati, C.~Song, and V.~Shmatikov, ``Membership inference
  attacks against machine learning models,'' in \emph{2017 IEEE Symposium on
  Security and Privacy (SP)}, San Jose, CA, USA, 2017, pp. 3--18.

\bibitem{b2}
J.~Alzubi, A.~Nayyar, and A.~Kumar, ``Machine learning from theory to
  algorithms: An overview,'' in \emph{Journal of Physics: Conference Series},
  vol. 1142, 2018.

\bibitem{b12}
\BIBentryALTinterwordspacing
Finnhub, ``Free realtime apis for stock, forex and cryptocurrency,'' Finnhub
  Stock APIs - Real-time stock prices, Company fundamentals, Estimates, and
  Alternative data, 2023, accessed: 17-Apr-2023. [Online]. Available:
  \url{https://finnhub.io/}
\BIBentrySTDinterwordspacing

\bibitem{b14}
\BIBentryALTinterwordspacing
T.~Srivastava, ``12 important model evaluation metrics for machine learning
  everyone should know (updated 2023),'' Analytics Vidhya, 2023, accessed:
  17-Apr-2023. [Online]. Available:
  \url{https://www.analyticsvidhya.com/blog/2019/08/11-important-model-evaluation-error-metrics/}
\BIBentrySTDinterwordspacing

\bibitem{b16}
S.~Kumari, K.~Singh, T.~Khan, M.~M. Ariffin, S.~K. Mohan, D.~Baleanu, and
  A.~Ahmadian, ``A novel approach for continuous authentication of mobile users
  using reduce feature elimination (rfe): A machine learning approach,''
  \emph{Mobile Networks and Applications}, Feb. 2023.

\bibitem{b6}
S.~Han, C.~Qubo, and H.~Meng, ``Parameter selection in svm with rbf kernel
  function,'' in \emph{World Automation Congress 2012}, Puerto Vallarta,
  Mexico, 2012, pp. 1--4.

\end{thebibliography}
\end{document}